\documentclass[10pt,twocolumn,letterpaper]{article}

\usepackage[pagenumbers]{cvpr} 

\usepackage[dvipsnames]{xcolor}

\definecolor{lightblue}{rgb}{0.1,0.3,1}
\definecolor{cvprblue}{rgb}{0.21,0.49,0.74}
\usepackage[pagebackref,breaklinks,colorlinks,citecolor=cvprblue]{hyperref}
\usepackage{dsfont}
\def\paperID{2656} 
\def\confName{CVPR}
\def\confYear{2024}

\title{Diffuse to Choose: Enriching Image Conditioned Inpainting in Latent Diffusion Models for Virtual Try-All}

\author{
Mehmet Saygin Seyfioglu\thanks{} \qquad Karim Bouyarmane\thanks{} \\ Suren Kumar \qquad Amir Tavanaei\qquad Ismail B. Tutar 
\\
\\
Amazon
\\
\\
\url{https://diffuse2choose.github.io}
}
\begin{document}

\twocolumn[{%
\renewcommand\twocolumn[1][]{#1}%
\maketitle
\begin{center}
    \centering
    \captionsetup{type=figure}
    \includegraphics[width=1\textwidth]{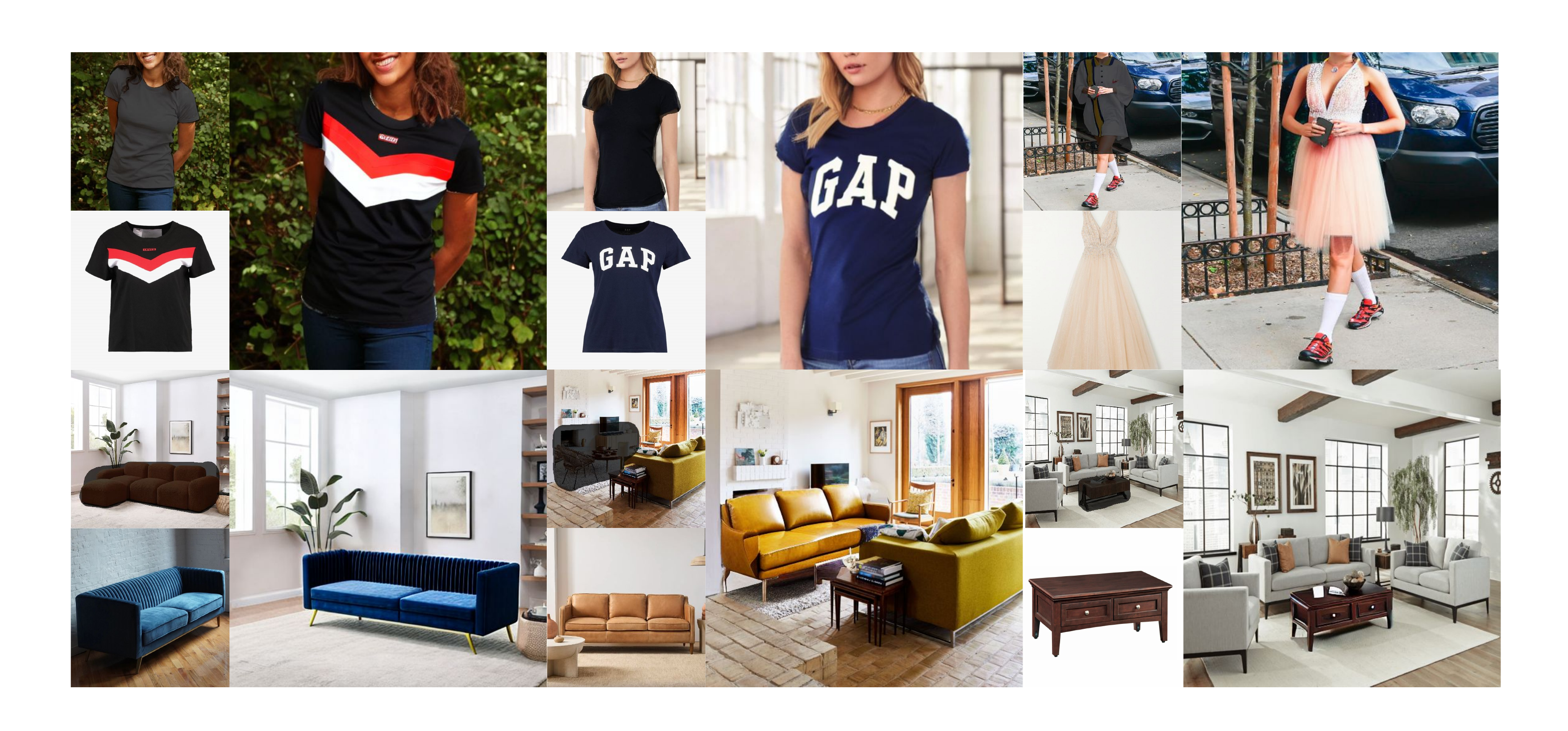}
    \captionof{figure}{Diffuse to Choose (DTC) allows users to virtually place any e-commerce item in any setting, ensuring detailed, semantically coherent blending with realistic lighting and shadows.}
    \label{fig:teaser}
\end{center}%
}]

{
  \renewcommand{\thefootnote}%
    {\fnsymbol{footnote}}
  \footnotetext[1]{University of Washington, work done during an internship at Amazon.}
  \footnotetext[2]{Correspondence at Amazon: bouykari@amazon.com}
}
\begin{abstract}


As online shopping is growing, the ability for buyers to virtually visualize products in their settings—a phenomenon we define as ``Virtual Try-All''—has become crucial. Recent diffusion models inherently contain a world model, rendering them suitable for this task within an inpainting context. However, traditional image-conditioned diffusion models often fail to capture the fine-grained details of products. In contrast, personalization-driven models such as DreamPaint are good at preserving the item's details but they are not optimized for real-time applications. We present "Diffuse to Choose," a novel diffusion-based image-conditioned inpainting model that efficiently balances fast inference with the retention of high-fidelity details in a given reference item while ensuring accurate semantic manipulations in the given scene content. Our approach is based on incorporating fine-grained features from the reference image directly into the latent feature maps of the main diffusion model, alongside with a perceptual loss to further preserve the reference item's details. We conduct extensive testing on both in-house and publicly available datasets, and show that Diffuse to Choose is superior to existing zero-shot diffusion inpainting methods as well as few-shot diffusion personalization algorithms like DreamPaint.

\end{abstract}    
\vspace{-2mm}
\section{Introduction}
\label{sec:intro}

The ever-growing demand for online shopping underscores the need for a more immersive shopping experience, allowing shoppers to virtually `try' any product from any category (clothes, shoes, furniture, decoration, etc.) within their personal environments. The concept of a \textbf{Virtual Try-All (Vit-All)} model hinges on its functionality as an advanced semantic image composition tool. In practice, this involves taking an image from a user, selecting a region within that image, and using a reference product image from an online catalog to semantically insert the product into the selected area while preserving its details. For such a model to be effective, it must fulfill three primary conditions: 1) operate in any 'in-the-wild' user image (not only on staged studios or professional human model images with predefined poses), and reference image, 2) integrate the reference product harmoniously with the surrounding context while maintaining the product's identity (not replacing the product with a generic image of a product from similar category), and 3) perform fast inference to facilitate real-time usage across billions of products and millions of users.

Existing solutions tend to be specialized. Instead of aiming for a general purpose Vit-All approach, models are often developed for specific tasks and domains (model for clothing, model for furniture, model for eyeglasses, etc.). For example, early GAN-based works focused primarily on virtual try-on of clothing on human models in limited contexts or controlled environment (such as only certain clothing segments and no in-the-wild user images, or product images) ~\cite{choi2021viton, xue2022dccf, bai2022single, lee2022high, lewis2021tryongan, karras2020analyzing}. Other approaches to the problem utilize somewhat expensive 3D AR/VR technologies for items like furniture in rooms~\cite{shum2023conditional, aubry2014seeing}, which is hard to scale to every single item on catalogs of billions of products that often lack 3D models. Consequently, a unified model offering a comprehensive Vit-All experience — one that enables consumers to digitally interact with any product from any category in any setting — is currently not available.

The emergence of diffusion models has marked a significant breakthrough in the generative capabilities of complex image modeling \cite{song2020score, rombach2022high, saharia2022photorealistic}. Unlike GANs, Diffusion models inherently grasp the nuances of the 3D world, exhibiting a degree of geometry and physics awareness, as demonstrated in inpainting tasks by \cite{zhan2023does}, establishing their usefulness for Vit-All applications. A DreamBooth-based~\cite{ruiz2023dreambooth} technique, called \textbf{DreamPaint \cite{seyfioglu2023dreampaint}}, showed that Stable Diffusion \cite{rombach2022high} can be few-shot fine-tuned for the Vit-All use case. It can infer how to warp clothes to a person's body, or how to place a certain furniture on a certain spot in a semantically correct manner without being explicitly trained to do so. While DreamPaint meets the first two criteria to be an effective Vit-All model, it requires few-shot fine-tuning for each product separately, compromising its suitability for real-time applications thus failing to meet the third criterion.

A recently introduced image-referenced inpainting model, Paint By Example (PBE)~\cite{yang2023paint}, operates in a zero-shot setting and can handle in-the-wild images, meeting criteria one and three. However it encounters a limitation due to its reliance on an information bottleneck in its conditioning process, utilizing only the [CLS] token of the reference image. This constraint leads to an over-generalization of the reference image, degrading the model's ability to maintain the fine-grained details essential for the Vit-All context, thus PBE fails to meet criterion two. Additionally, operating within a latent space, PBE struggles to retain fine-grained details of each item, underscoring the necessity for incorporating some form of pixel-level guidance.

\begin{figure*}[h!]
    \centering
    \includegraphics[width=0.75\textwidth]{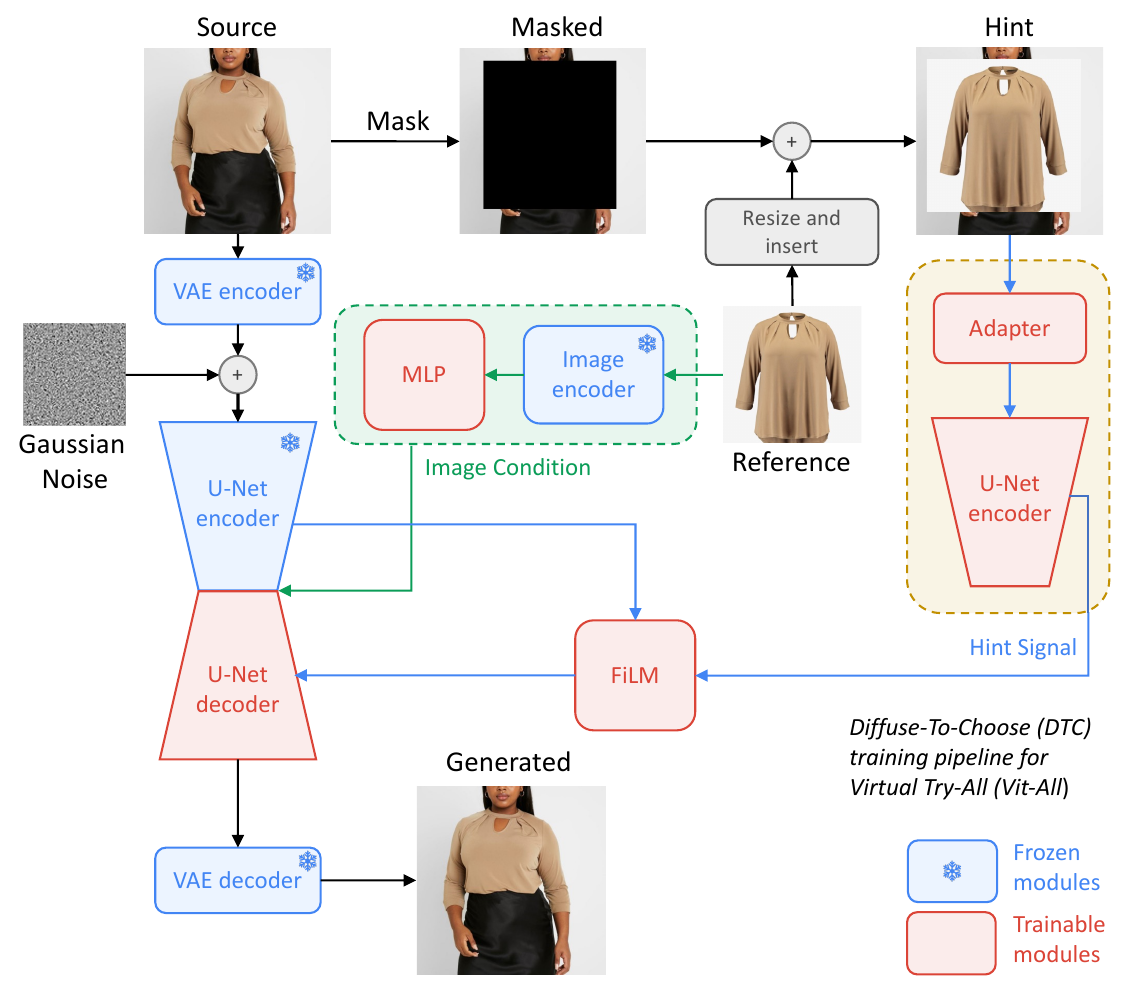}
    \caption{Pipeline of Diffuse to Choose. The process initiates with masking the source image, followed by inserting the reference image within the masked area. This pixel-level `hint' is then adapted by a shallow CNN to align with the VAE output dimensions of the source image. Subsequently, a U-Net Encoder processes the adapted hint. Then, at each U-Net scale, a FiLM module affinely aligns the skip-connected features from the main U-Net Encoder and pixel-level features from the hint U-Net Encoder. Finally, these aligned feature maps, in conjunction with the main image conditioning, facilitate the inpainting of the masked region. \textcolor{red}{Red} indicates trainable modules and \textcolor{lightblue}{blue} indicates frozen modules. }
    \label{fig:dtc_pipeline}
\end{figure*}

\begin{figure*}[h]
\centering
\includegraphics[width=1\textwidth]{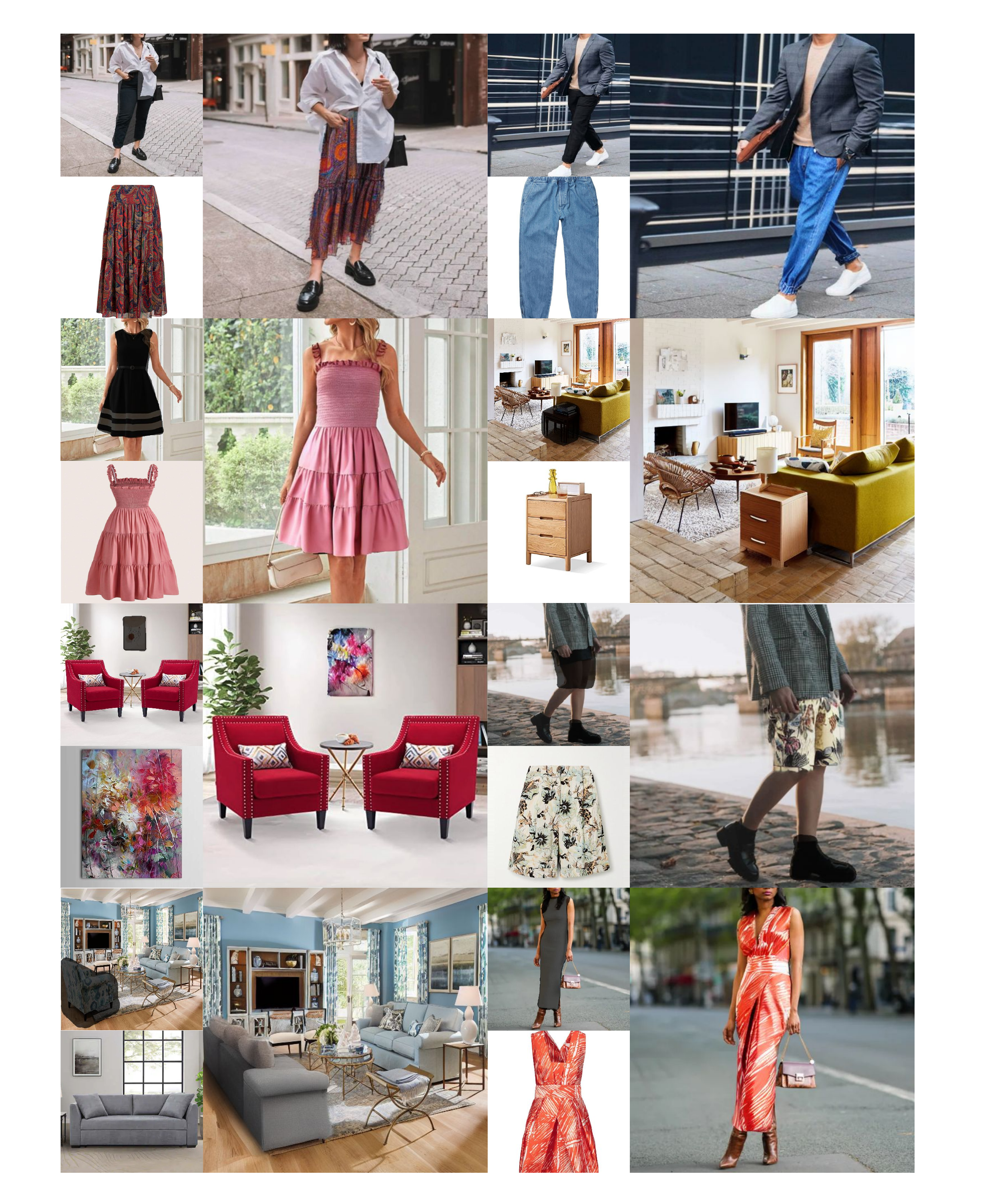}
\caption{DTC can handle variety of e-commerce products and can generate images using in-the-wild images \& references.  }
\label{fig:dct_large_results}
\end{figure*}

In this work, we introduce \textbf{``Diffuse to Choose'' (DTC)}, a novel diffusion inpainting approach designed for the Vit-All application. DTC, a latent diffusion model, effectively incorporates fine-grained cues from the reference image into the main U-Net decoder using a secondary U-Net encoder. Inspired by ControlNet \cite{zhang2023adding}, we integrate a pixel-level ``hint'' into the masked region of an empty image, which is then processed through a shallow convolutional network, ensuring dimensional alignment with the masked image processed by the Variational Autoencoder (VAE). DTC harmoniously blends the source and reference images, maintaining the integrity and details of the reference image. To further enhance alignment of basic features such as color, we employ perceptual loss using a pre-trained VGG model \cite{gou2023taming}. The complete architecture is illustrated in Fig.~\ref{fig:dtc_pipeline}, with examples showcased in Fig.~\ref{fig:teaser} and Fig.~\ref{fig:dct_large_results}.

DTC effectively fulfills all three criteria for the Vit-All use case: 1) It efficiently handles in-the-wild images and references, 2) It adeptly preserves the fine-grained details of products while ensuring their seamless integration into the scene, and 3) It facilitates rapid zero-shot inference. We trained DTC on an in-house training dataset with sampled 1.2M source-reference pairs and a smaller public dataset, VITON-HD-NoFace \cite{choi2021viton}\footnote{The VITON-HD public dataset was modified to remove and crop all model faces from the images in the dataset.}. Our quantitative evaluations and human studies demonstrate that DTC surpasses all PBE variants — for which we implemented several enhancements to facilitate a fair comparison against DTC — and matches the performance of non-real-time, few-shot personalization models like DreamPaint \cite{seyfioglu2023dreampaint}, within the Vit-All context.

\section{Related Work}
\noindent
\textbf{Virtual Try-On}. The primary goal of virtual try-on approaches is to create an image of a person wearing a target garment, ensuring that the garment's fine-grained details are preserved and that it blends seamlessly with its surrounding context. Since this is a highly domain-specific and constrained problem (models and garments are often not presented in in-the-wild examples), existing models generally employ warp and paste (blend) techniques, leveraging extra inputs such as pose and human parse maps~\cite{choi2021viton, xue2022dccf, bai2022single, lee2022high, lewis2021tryongan, karras2020analyzing, gou2023taming, xie2023gp, yan2023linking}. VITON \cite{han2018viton} uses a two-step synthesis and refinement process, initially generating a coarse image with the desired clothing and then refining it to enhance the details. VITON-HD \cite{choi2021viton} aims for higher-resolution images and employs alignment-aware segment normalization to correct misaligned parts. TryOnGAN \cite{lewis2021tryongan} uses pose conditioning but relies on a purely latent model, which often lacks fine-grained details when representing garments. Given that GAN-based approaches are tailored exclusively for virtual try-on, which is a more narrowly defined problem compared to ours, and they do not possess the broad mode coverage inherent to diffusion models, we have chosen to concentrate solely on diffusion models in our work. One of the latest diffusion-based virtual try-on models, Tryondiffusion \cite{zhu2023tryondiffusion}, employs a dual U-Net approach on a pixel-level diffusion model. While it offers impressive performance for virtual try-on, it struggles with in-the-wild examples, supports only upper garments, and is not suitable for real-time use. Thus, a latent diffusion approach is necessary to ensure real-time inference in practical use cases. Regarding furniture, most existing works are AR-based 3D approaches from large corporations, which do not provide much detail about their models. In the 2D domain, there is DreamPaint \cite{seyfioglu2023dreampaint}, a DreamBooth-based \cite{ruiz2023dreambooth} inpainting approach that few-shot fine-tunes the U-Net of Stable Diffusion, but it is not suitable for real-time applications.

\noindent
\textbf{Diffusion Based Image Editing}. Creating realistic composites by superimposing an object from one image onto another is a common practice in photo editing and is closely aligned with the Vit-All task. Image Editing, particularly inpainting, has been extensively explored in diffusion models. Initially, there were text-based image editing models \cite{rombach2022high, avrahami2022blended, hertz2022prompt, kawar2023imagic, kim2022diffusionclip}. However, it is evident that text alone cannot capture the fine-grained details necessary for accurately describing a product, necessitating the use of image conditioning. DCCF \cite{xue2022dccf} introduced pyramid filters for image composition, followed by Paint by Example \cite{yang2023paint}, which conditions the diffusion model using CLIP embeddings of the reference image. However, relying solely on the [CLS] token often leads to an overgeneralization of the reference image, making it unsuitable for the Vit-All task. Similarly, ObjectStitch \cite{song2023objectstitch} combines image and text embeddings to guide the model but faces challenges in conveying fine-grained details. In response to these challenges, we introduce Diffuse to Choose, a novel latent diffusion inpainting model. Our model effectively leverages fine-grained details from the reference image, ensuring both the preservation of the product's fine-grained details and its seamless integration into the chosen location, while working in a zero-shot setting with any in-the-wild image.

\section{Method}

\begin{figure}[t!]
    \centering
    \includegraphics[width=0.8\linewidth]{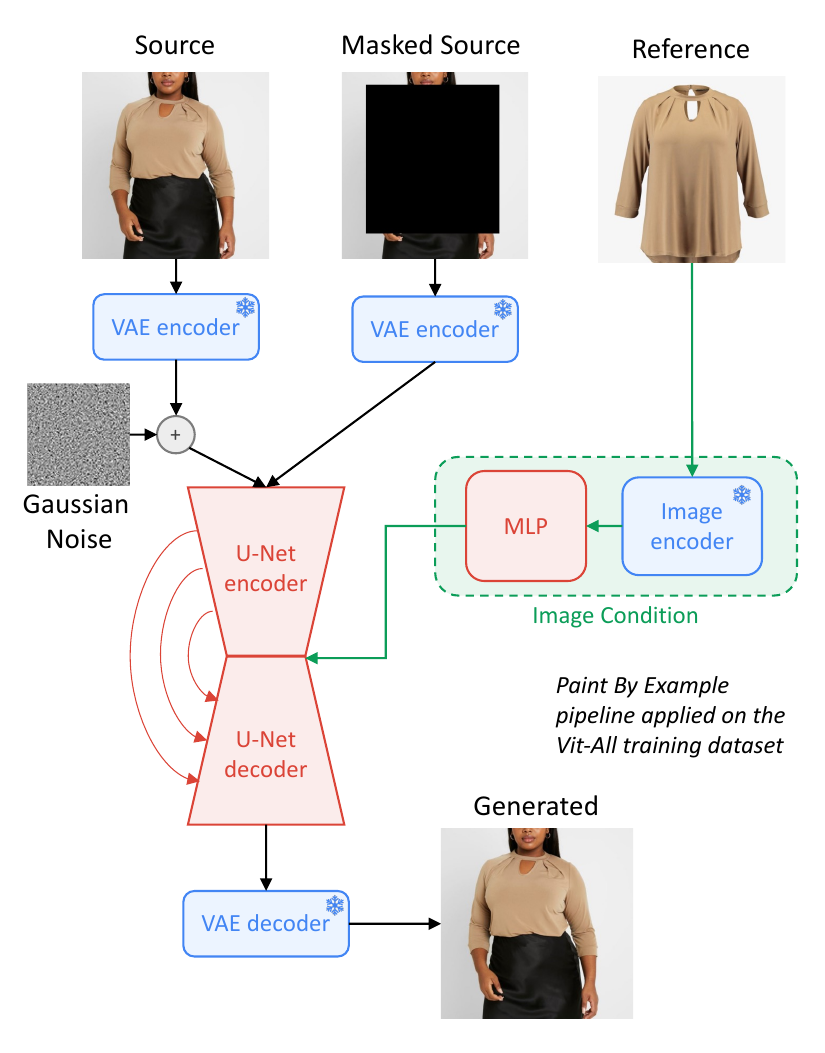}
    \caption{Pipeline of Paint by Example \cite{yang2023paint} for Vit-All case. \textcolor{red}{Red} are trainable and \textcolor{lightblue}{blue} are frozen. Orange pathways indicate skip connections between the encoder and the decoder.}
    \label{fig:pbe_pipeline}
\end{figure}

We formulate Vit-All as an image-conditioned inpainting task, wherein a single product image is integrated into a user-specified region within a user-specified image, ensuring the preservation of the product's fine-grained details and its harmonious blend with the target image. A naive approach would be using the conventional PBE method, shown in Fig. \ref{fig:pbe_pipeline}. However, due to the information bottleneck caused by PBE's reliance on only the [CLS] token for image conditioning, it tends to lose significant details of the reference image, resulting in unsatisfactory performance.

To rectify the shortcomings of the PBE in preserving the reference image's details, we introduce the ``Diffuse to Choose'' (DTC) method. DTC leverages an auxiliary U-Net alongside the primary U-Net within a latent diffusion model, specifically Stable Diffusion v1.5 \cite{rombach2022high}. The purpose of the auxiliary U-Net is to protect the details of the reference image that might be lost due to both the latent nature of the Stable Diffusion model and the limitations of image conditioning. To this end, we directly infuse fine-grained details of the reference image into the main U-Net's decoder via affine transformations, ensuring preservation of the reference product's details in the generated image. Our pipeline is shown in Fig. \ref{fig:dtc_pipeline}.

\subsection{Diffusion Inpainting Models}

For the Vit-All inpainting task, our objective is as follows: Given a user provided source image, $\mathbf{x}_s\in \mathbb{R}^{H\times W\times 3}$, and a user-defined mask, $m$, (is a binary matrix of dimension ${0,1}^{H x W}$), with zeros indicating editable regions, and a reference image $x_r$, showcasing the desired product, the objective is to seamlessly incorporate the product image ${x}_r$ within the mask-defined region of ${x}_s$, ensuring the preservation of ${x}_r$'s details. Diffusion models provide unparalleled success in image generation and specific tasks such as inpainting ~\cite{sohl2015deep, ramesh2021zero, saharia2022photorealistic, rombach2022high}. These models follow a Markovian process, gradually introducing noise, denoted as $\epsilon$ from $\mathcal{N}(0,1)$, to $x_s$ over various timesteps $t$ to transform it into an isotropic Gaussian distribution $z_t$. The process is then reversed by iteratively predicting and subtracting the added noise to convert $z_t$ back to $x_s$, conditioned by $c$. In the context of inpainting, this can be mathematically expressed as:

\begin{equation}
    L=\mathbb{E}_{z_t,\epsilon, t}\left[\left\|\epsilon_\theta\left((m \odot x_s), z_t, t, c\right) - \epsilon\right\|_2^2\right]
\label{eq:inpaint}
\end{equation}

Here, $x_s$ is the source image, $m$ the user-defined mask, $z_t$ the noise-added version of $x_s$, and $c$ denotes the embeddings of $x_r$. PBE uses the [CLS] token of CLIP \cite{radford2021learning} for $c$, a deliberate information bottleneck, because it relies on self-referencing, and additional patches often leads to copy-paste artifacts. However, for the Vit-All task, it is practical to compile a dataset with distinct source and reference images of the same object, thereby eliminating this bottleneck. Consequently, we introduced a series of enhancements to PBE to explore the upper limits of basic image-conditioned inpainting models in the Vit-All context and establish a stronger baseline. Our modifications included using all CLIP patches instead of just the [CLS] token, employing a larger image encoder DINOV2 \cite{oquab2023dinov2}, and adding a refinement loss similar to \cite{gou2023taming} alongside the diffusion loss given in Eq. \ref{eq:inpaint}. Each of these alterations incrementally improved the performance of the PBE approach.

\subsection{Design of Diffuse to Choose}

\begin{figure}[t!]
    \centering
    \includegraphics[width=0.8\linewidth]{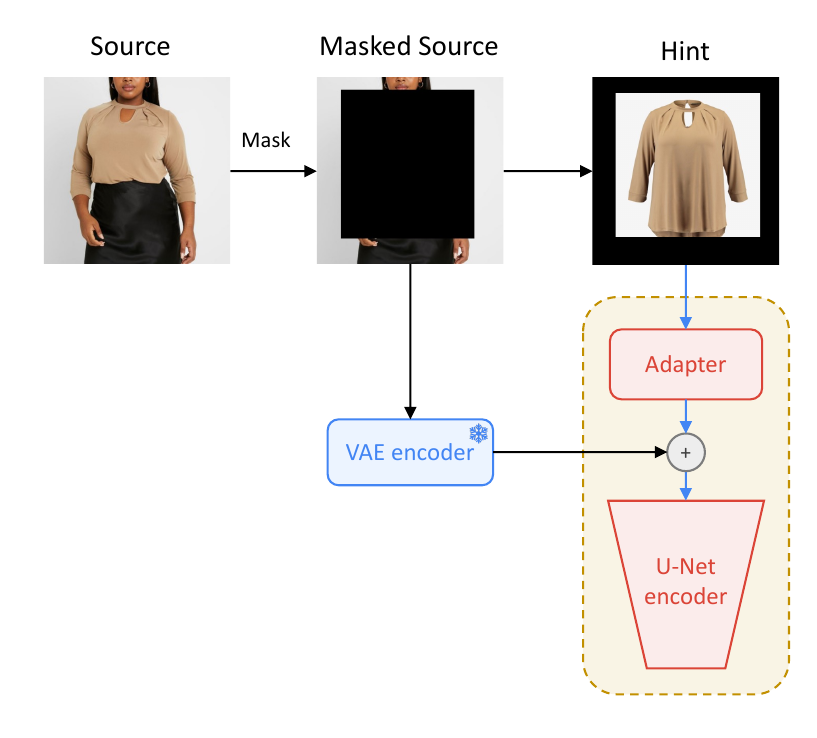}
    \caption{Hint signal is stitched into a blank image within the masked region, then summed up with latent masked input before fed into Auxiliary U-Net.}
    \label{fig:hintpathway}
\end{figure}

\noindent
\textbf{Creating the Hint Signal}. Drawing inspiration from ControlNet \cite{zhang2023adding}, we propose the incorporation of a secondary U-Net encoder, which serves as a trainable replica of the main U-Net encoder. In the ControlNet architecture however, the secondary U-Net is integrated directly into the main U-Net decoder, providing spatial conditioning. In contrast, DTC demonstrates that the secondary U-Net, rather than providing a spatial layout, can serve to guide the main U-Net by exerting a potent pixel-wise influence from the reference image during the decoding process. To generate the hint signal, we start by creating an image of zeros, identical in size to the source image, $\mathbf{x}_s\in \mathbb{R}^{H\times W\times 3}$. Subsequently, we resize the reference image and insert it within the designated mask coordinates within the image of zeros. The same mask is then applied to $x_s$, and this masked source image undergoes processing by the VAE encoder to yield a latent representation, sized $64\times64\times4$. The Hint image is subsequently processed by the Adapter module —a shallow convolutional network comprising four layers— to match with the dimensions of the masked latent inpaint image. Finally, the Hint image and the masked source are added elementwise to produce the final representation of the hint image, which is then processed by the replicated U-Net encoder. This process is not shown in Fig. \ref{fig:dtc_pipeline} to keep it concise, but is illustrated in Fig. \ref{fig:hintpathway} for clarity. Through a series of ablation studies, we demonstrate that maintaining a distinct representation for the hint image at the pixel level, while keeping the inpaint image in a latent form, provides complementary signals that yield superior results.

\noindent
\textbf{Combining Hint Signal with Main U-Net}. The Stable Diffusion U-Net Encoder generates feature maps of varying resolutions at each level, consisting of 13 blocks, including the middle layer. The direct addition of the Hint Encoder's outputs to the skip connections of the Main U-Net Encoder at every level tends to result in a pronounced spatial influence from the Hint signal, which is often not spatially aligned with the source image, thus negatively affecting the performance. In addition to direct addition, we explore two strategies for integrating the Hint Signal into the main U-Net: Using Feature-wise Linear Modulation (FiLM) \cite{perez2018film}, and Cross Attention, computed in a manner akin to \cite{zhu2023tryondiffusion}. Among these three approaches—direct addition, FiLM, and Cross Attention—FiLM emerges as the most effective. We argue this is due to the image conditioning already capturing the majority of low-level details from the reference image, with mostly fine-grained details being absent. FiLM specifically enhances those feature channels that are essential for preserving the fine-grained details of the reference image.

\noindent
\textbf{Hinting Strategies, Refinement Loss and Image Encoder.} Our objective is to convey pixel-level, fine-grained details from the reference image into the main U-Net, and there are several methods to achieve this. One approach is to focus on high-frequency details by employing techniques like Canny Edges or Holistically-Nested Edge Detection (HED) features. Alternatively, we can directly use the reference image itself. In our experiments, we tested Canny edge extraction using the implementation in the OpenCV library \cite{opencv_library}, with minimum and maximum threshold values of 30 and 140, and for slightly softer edges, we used the HED model \cite{xie2015holistically}. Despite these strategies yielding comparable results, directly using the reference image proved to be the most effective as it conveys the entire spectrum of details from the reference image, rather than focusing solely on high-frequency details. Thus, instead of pre-filtering to only convey the high-frequency details, it is a better approach to let the FiLM layer decide the most important channels, thus capturing the essential nuances of the reference image.

For our image encoder, we use DINOV2 \cite{oquab2023dinov2}, which outputs $256\times1536$ dimensional embeddings to represent a reference image, which is then reduced to $256\times768$ by a trainable MLP layer. Finally, we utilized a perceptual loss using a pre-trained VGGNet \cite{simonyan2014very}, which is computed by comparing the feature maps from the first five layers of VGGNet for both the source and the generated images, thereby implicitly ensuring the alignment of basic features like color.

\section{Experiments}
\subsection{Dataset and Implementation Details}

\textbf{Data.} We compiled an in-house training dataset composed of product images. Fortunately, e-commerce products often have multiple images available, so, during training, we do not need to adhere to the self-reference setting employed by PBE, where the reference image is derived from $x_s$, leading to potential overfitting. However not all products yield useful $x_s, x_r$ pairs, as many product images feature only the product against a white background.  While these images are apt for use as $x_r$, they are unsuitable for $x_s$ since we require images of products in contextual settings (with a natural background). To address this, we employed an in-house model to identify products that have at least one $x_s$ image depicting the product in a natural setting, interacting with other elements in the scene, and one image with the product itself $x_r$, where we collect images against a white background if it exists. Finally, we use GroundingDINO \cite{liu2023grounding} and SAM \cite{kirillov2023segment} alongside with the product type of the $x_s$ to create the inpainting mask within $x_s$. From the resulting dataset, we sampled a sample training dataset of 1.2M samples, evenly split between wearables and furniture. To ensure accessibility and reproducibility, we also train and test our model on a public dataset modified to remove model faces, VITONHD-NoFace \cite{choi2021viton}, which provides $x_r$ against a white background, masks, and $x_s$ where individuals (with removed faces) are wearing $x_r$.

\noindent
\textbf{Implementation details.} We use a latent diffusion model, Stable Diffusion \cite{rombach2022high} V1.5 as our backbone in our experiments. Our image resolution is $512 \times 512$ and we train our model with DDPM \cite{ho2020denoising} using a constant learning rate of $1e-5$ in both PBE and DTC implementations. We use simple augmentations like rotation and flip but avoid strong augmentations given in \cite{yang2023paint}, as we don't rely on self-referencing. We also use classifier free guidance \cite{ho2022classifier} in similar fashion to \cite{zhang2023adding}. During inference, we use DDIM \cite{song2020denoising} with guidance scale of 5, and for the hint input we stitch the reference image into the largest rectangular bounding box within the arbitrary-shape binary mask. We use 8 NVIDIA A100 40G GPUs to train our model for 40 epochs.

\subsection{Paint by Example Ablations}
\begin{figure}[h!]
    \centering
    \includegraphics[width=\linewidth]{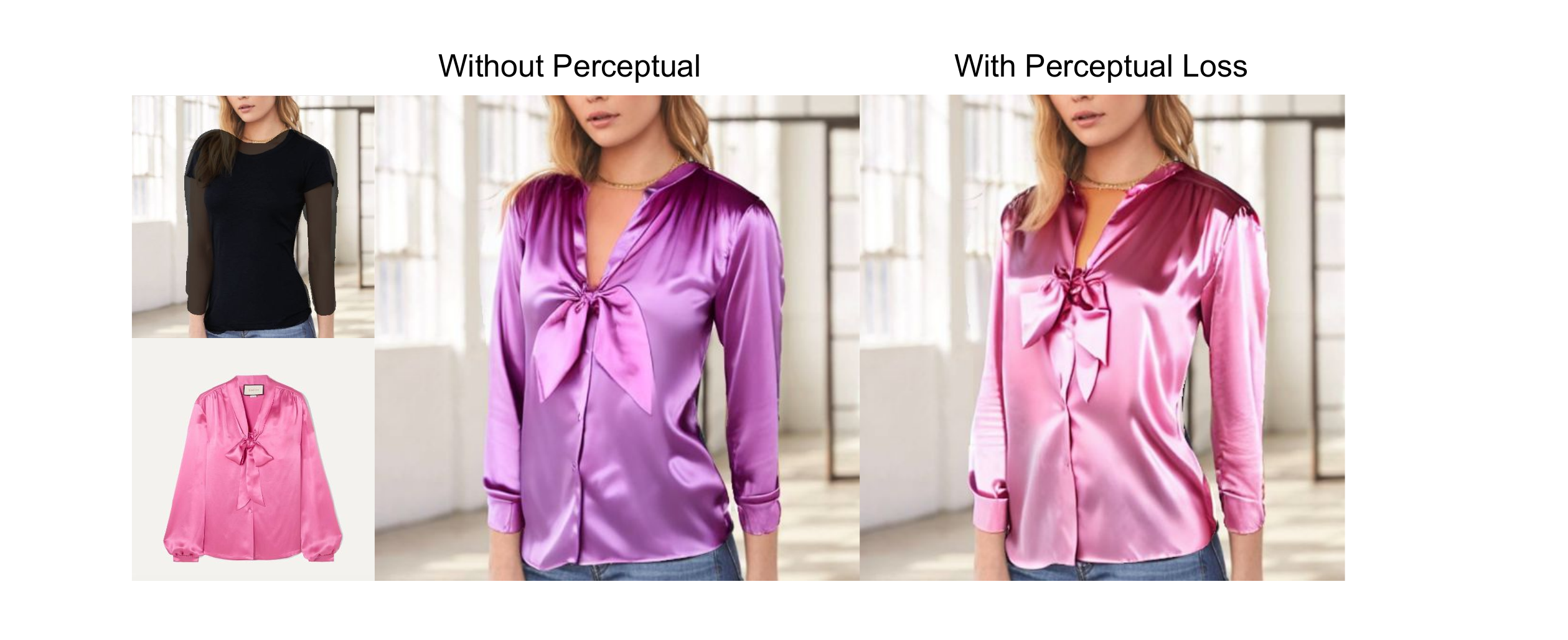}
    \caption{Effect of Perceptual Loss.}
    \label{fig:perceptual}
\end{figure}

To ascertain the optimal performance of the naive image-conditioned inpainting approach \cite{yang2023paint} and create the strongest possible baseline, we implemented a series of modifications to the architecture illustrated in Fig.~\ref{fig:pbe_pipeline}. Originally, PBE utilized self-reference conditioning, which involved cropping the $x_r$ from within $x_s$. However, in the Vit-All context, we circumvent this limitation by having separate images for $x_r$. As a result, instead of using only the [CLS] token from CLIP \cite{radford2021learning} in the Image Encoder, we incorporated all CLIP patches alongside the [CLS] token. Subsequently, we increased the capacity of the image encoder by adopting DINOV2 \cite{oquab2023dinov2}, a larger and purely image-based model. Furthermore, akin to \cite{gou2023taming}, we integrated a perceptual loss with the diffusion loss. This involved using an ImageNet pre-trained VGGNet \cite{simonyan2014very} to foster alignment of basic features, such as color and certain textures, between the generated and the source images. A qualitative example of this implementation is presented in Fig.~\ref{fig:perceptual}.

\subsection{Diffuse-to-Choose Ablations}

\noindent
\textbf{Hint Pathway Ablations.} It is possible to directly insert the reference image $x_r$ into the masked source image $m \odot x_s$ and process it with a VAE, circumventing the use of the adapter network to align the hint image's resolution with the latent masked image prior to addition. However, this approach produce suboptimal results, yielding a Clip Score of $86.97$ and an FID of $6.26$ on our Vit-All dataset, in contrast to the non-latent Hint insertion's Clip Score of $88.14$ and FID of $5.72$. We hypothesize that maintaining the hint signal at the pixel level could introduce additional information that is overlooked in its VAE encoded latent counterpart. This indicates that the VAE might be excluding certain features during the encoding process.

\noindent
\textbf{Ablations on Hinting Strategies.} We explored alternatives to directly inserting the reference image into the masked source image. These alternatives included using Canny Edges and HED features, both of which are designed to convey the high-frequency details that are absent in image-only conditioning. However, we observed a slight underperformance with both HED and Canny edges compared to the direct use of the reference image. This was evidenced by the CLIP scores, which were $87.85$ for Canny and $86.98$ for HED, compared to $88.14$ for direct usage on our Vit-All dataset. Similarly, the FID scores were $6.11$ for Canny and $6.57$ for HED, against $5.72$ for direct insertion.


\begin{table}[h!]
\caption{Quantitative comparison between DTC variants and $\text{PBE}_{\text{best}}$, which denotes a PBE variant using DINOv2 and perceptual loss. CA denotes Cross-Attention.}
\vspace{-0.2cm}
\small
\centering
\begin{tabular}{@{}lcccc@{}}
\toprule
Method  & CLIP Score  ($\uparrow$) & FID ($\downarrow$) \\ \midrule
$\text{PBE}_{\text{best}}$    & 85.43  & 6.65    \\
$\text{Ours}_{\text{addition}}$ & 86.94  & 6.19   \\
$\text{Ours}_{\text{CA}}$ & 88.01  & \textbf{5.68}   \\
$\text{Ours}_{\text{FiLM}}$ & \textbf{88.14}  & 5.72   \\
\bottomrule
\end{tabular}
\vspace{-0.4cm}
\label{tab:final_results}
\end{table}

\noindent
\textbf{Ablations of Techniques for Integrating Hint Signal into the Main U-Net.} There are multiple ways to merge the signals from the hint U-Net and the main U-Net before incorporating the combined signal into the main U-Net decoder. We explored three approaches: direct addition, the use of an affine transformation layers FiLM \cite{perez2018film}, and the integration of more computationally expensive Cross Attention layers \cite{zhu2023tryondiffusion}. Results shown on Tab. \ref{tab:final_results} revealed that both FiLM and Cross Attention layers outperform direct addition. Also, Cross Attention and FiLM yield comparable results, and FiLM is cheaper to compute, therefore we chose to use FiLM in our final model.

\begin{figure}[h!]
    \centering
    \includegraphics[width=\linewidth]{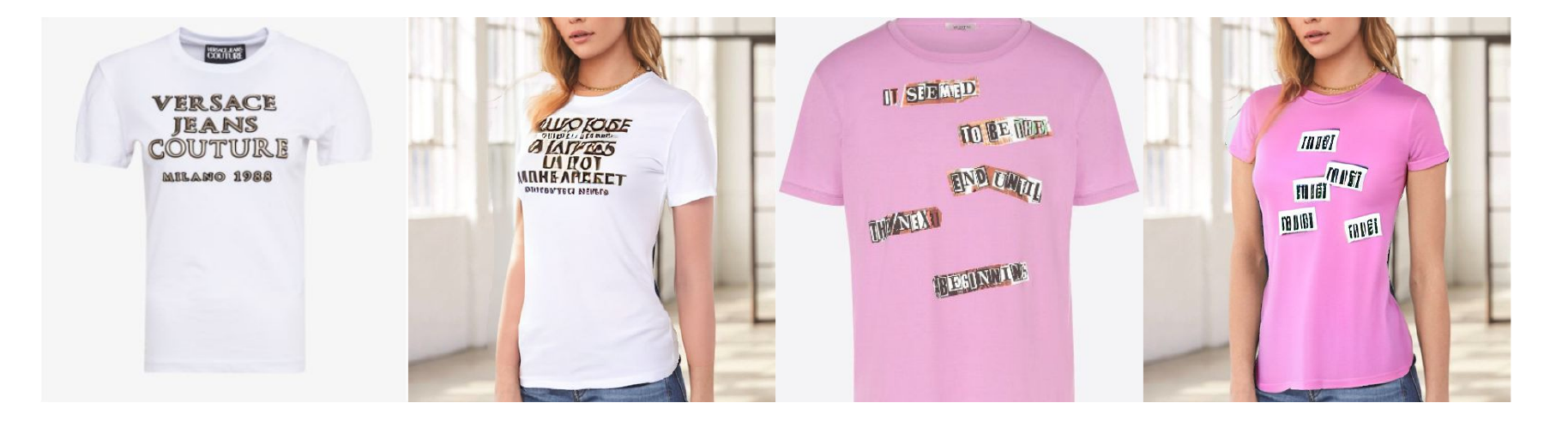}
    \caption{Failure cases with generating fine-grained text.}
    \label{fig:failure_cases}
\end{figure}

\begin{figure*}[t]
\centering
\includegraphics[width=1\textwidth]{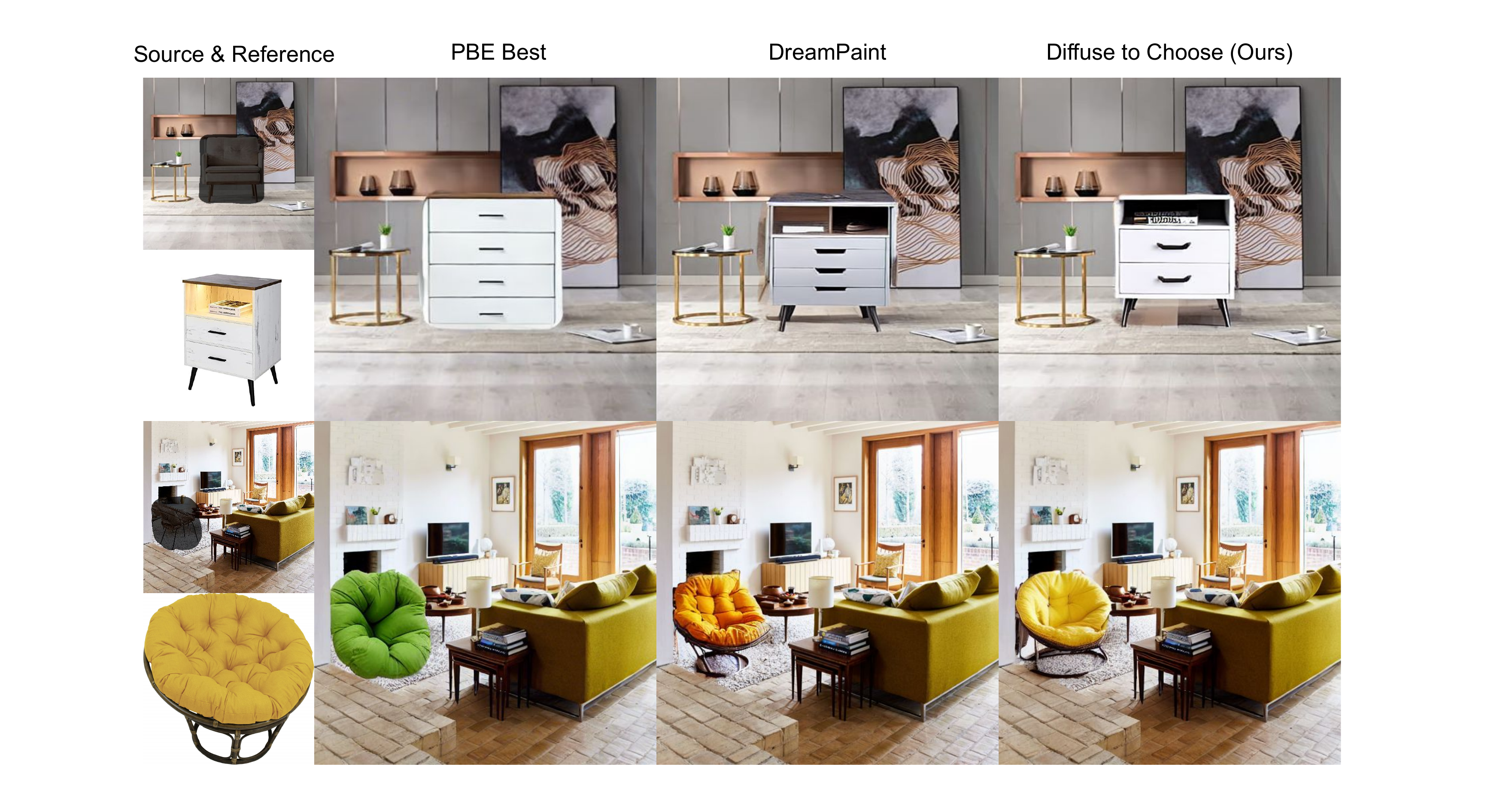}
\vspace{-0.6cm}
\caption{Qualitative comparison against PBE and DreamPaint.}
\vspace{-0.4cm}
\label{fig:dreampaint_comparisons}
\end{figure*}

\begin{figure*}[h!]
    \centering
    \includegraphics[width=\linewidth]{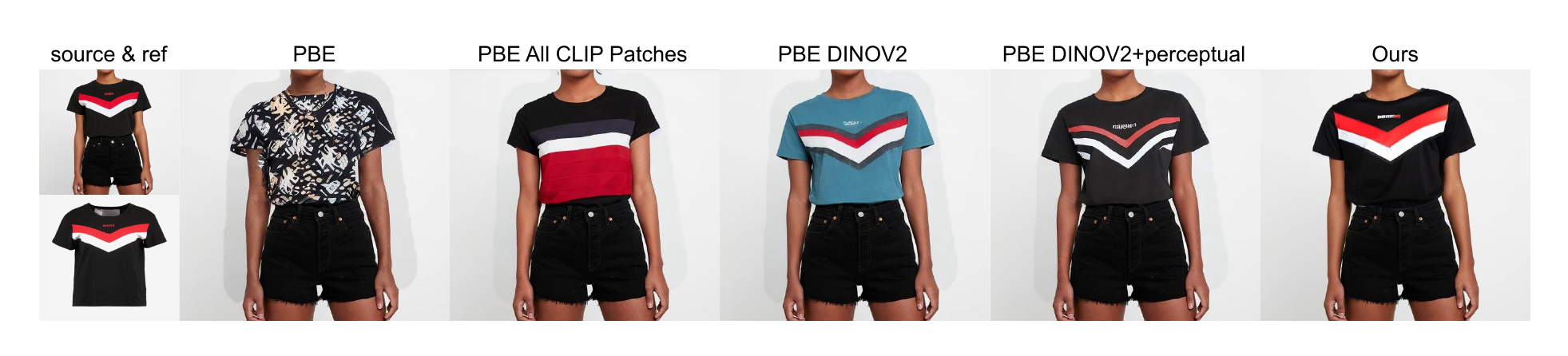}
    \caption{PBE variants' performance vs DTC}
    \label{fig:pbevariants}
\end{figure*}


\begin{table}[b]
\caption{Quantitative comparison of PBE variants on VITONHD-NoFace \cite{choi2021viton}.}
\vspace{-0.2cm}
\small
\centering
\begin{tabular}{@{}lcccc@{}}
\toprule
Method  & CLIP Score  ($\uparrow$) & FID ($\downarrow$) \\ \midrule

PBE CLIP$_{\text{cls}}$\cite{yang2023paint}    & 82.43  & 9.54    \\
+ PBE $\text{CLIP}_{\text{all}}$  & 84.01  & 8.93    \\ 
+ PBE DINOv2   & 87.48  & 6.18   \\
+ PBE perceptual   &  87.79 & 5.93  \\
\textbf{Ours} & \textbf{90.14}  & \textbf{5.39}   \\
\bottomrule
\end{tabular}
\vspace{-0.4cm}
\label{tab:quantitative_ablation_pbe}
\end{table}

\vspace{-0.2cm}
\subsection{Evaluation and Comparisons}

\noindent
\textbf{Comparison Against Paint by Example Variants.} 
 We implemented a series of enhancements to PBE and trained each variant on VITONHD-NoFace dataset. The results are presented in Table \ref{tab:quantitative_ablation_pbe}. As anticipated, using all CLIP patches surpasses the performance of using only the [CLS] token, which is limited to encoding a generalized version of $x_r$. Furthermore, augmenting the size of the image encoder by using DINOv2 notably enhances performance. Notably, the addition of perceptual loss provides a marginal improvement in scenarios where the model initially struggled with basic features, such as color. While PBE, particularly with DINOv2 and perceptual loss, is adept at handling basic items with minimal details, it often falls short in the inpainting of detailed items. In contrast, DTC exhibits superior performance, especially in preserving the fine-grained details of items. Figure \ref{fig:pbevariants} illustrates the outcomes achieved with certain enhancements.

\begin{table}[!h]
\caption{The average results of the small-scale human evaluation study. Similarity evaluates the resemblance of the inpainted region to the reference image, while Semantic Blending assesses the accuracy of the reference image's integration within its context.}
\vspace{-0.2cm}
\small
\centering
\begin{tabular}{@{}lcccc@{}}
\toprule
Method  & Similarity  ($\downarrow$) & Semantic Blending ($\downarrow$) \\ \midrule
$\text{PBE}_{\text{best}}$    & 3.7  & 3.13    \\
DreamPaint \cite{seyfioglu2023dreampaint}    & \textbf{2.83}  & 2.53    \\
\textbf{Ours} & 2.9  & \textbf{2.5}  \\
\bottomrule
\end{tabular}
\vspace{-0.4cm}
\label{tab:dreampaint_quant}
\end{table}

\noindent
\textbf{Comparisons Against Few-Shot Personalization Methods.} While personalization methods such as DreamBooth \cite{ruiz2023dreambooth} do not support inpainting, the recently introduced DreamPaint approach \cite{seyfioglu2023dreampaint} enables similar few-shot fine-tuning of the U-Net in a masked setting, allowing for the generation of specified concepts at user-defined locations. However, DreamPaint requires few-shot fine-tuning with multiple product images, taking about 40 minutes per product to be trained. We manually selected 30 samples to compare DTC with DreamPaint and PBE. Visual comparisons are presented in Fig. \ref{fig:dreampaint_comparisons}. Furthermore, we conducted a subjective human survey, the results of which are tabulated in Table \ref{tab:dreampaint_quant}. A total of 20 participants scored each image on a scale from 1 to 5, with 1 being the best, based on both the inpainted region's similarity to the reference image and its contextual blending. The results show that DTC, despite being a zero-shot model, performs on par with DreamPaint, which requires few-shot fine-tuning with multiple $x_r$.

\section{Conclusion and Limitations}

\noindent
\textbf{Limitations.} DTC has limitations. Despite our efforts to inject fine-grained details, the model may still overlook fine-grained details, particularly in text engravings, a challenge inherent to Stable Diffusion (see Fig. \ref{fig:failure_cases}). Additionally, the model might alter human poses since it doesn't consider pose, leading to discrepancies with pose-agnostic masking, especially for full-body coverage (see Fig. \ref{fig:poseagnosticmask} in the Appendix). Introducing pose conditioning could mitigate this, but we prioritized a general-purpose model over task-specific auxiliary inputs for broader applicability.

\noindent
\textbf{Conclusion.} In this paper, we introduced "Diffuse to Choose," a novel image-conditioned diffusion inpainting model designed for the Virtual Try-All, aiming to integrate e-commerce items into user images while preserving item details. Our main contributions include employing a secondary U-Net to infuse fine-grained signals from the reference image into the primary U-Net decoder using basic affine transformation layers within a latent diffusion model. Moreover, we refined the PBE model for peak performance achievable with straightforward image-conditioned inpainting models. We compared DTC with upgraded PBE variants and a few-shot personalization methods using datasets like VITONHD-NoFace and a larger in-house  dataset and show that DTC outperforms existing diffusion based inpainting approaches in the Virtual Try-All setting.

{
    \small
    \bibliographystyle{ieeenat_fullname}
    \bibliography{main}
}

\clearpage
\setcounter{page}{1}
\maketitlesupplementary

\subsection{Masking Strategy During Training and Inference}

\begin{figure*}[h!]
\begin{center}
  \includegraphics[width=1.0\linewidth]{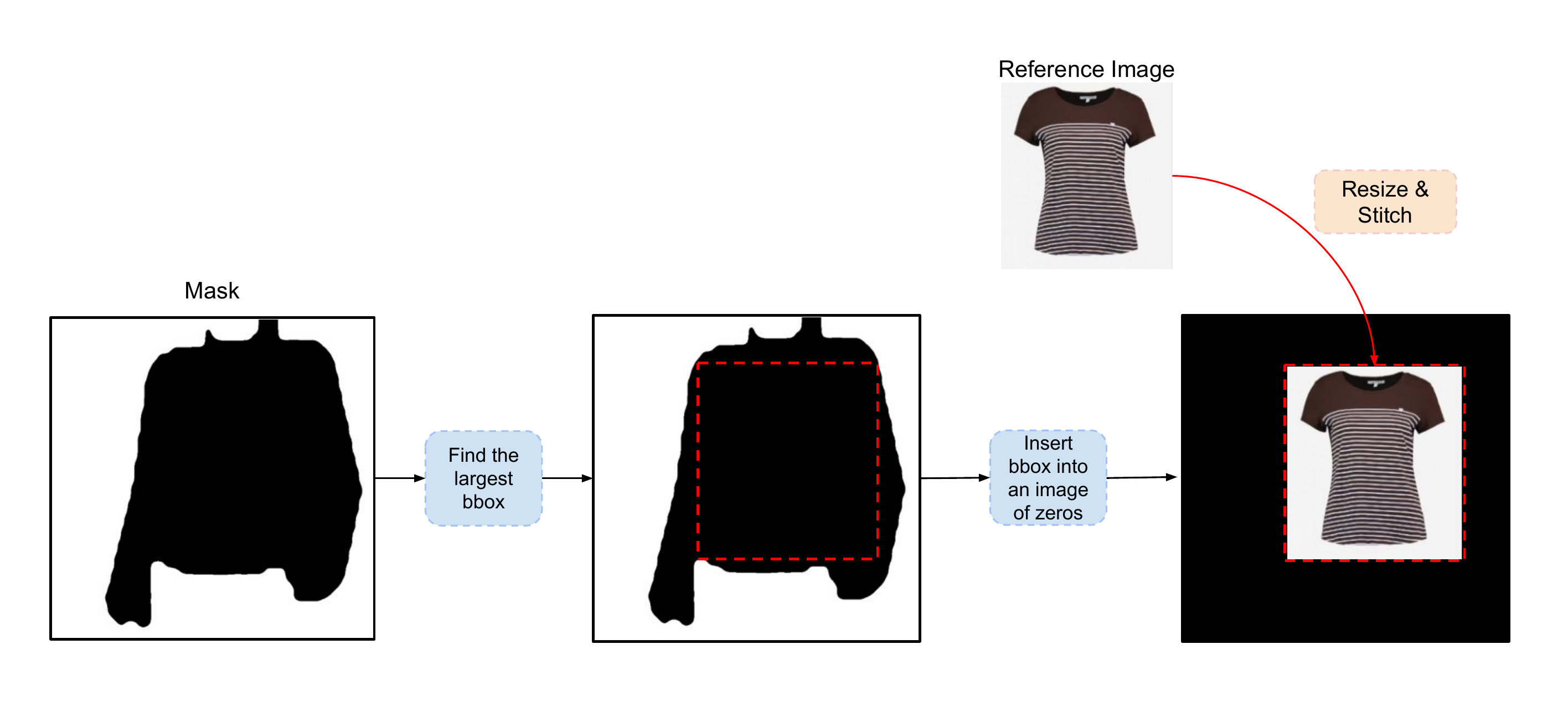}
\end{center}
  \caption{We find the largest rectangular bounding box inside an fine-grained binary mask. Then the same coordinates are used to stitch the reference image into an image of zeros to create the initial hint signal.}
\label{fig:mask_shape_example}
\end{figure*}

During training, with equal probability, we alternate between a fine-grained mask (where we only mask the item specifically) and bounding box shaped masks  (covering the largest bounding box spanned by the fine-grained mask). For each case, we stitch the reference image within the largest rectangular shape inside the mask. This approach is straightforward in the case of rectangular masks. However, for fine-grained masks, we calculate the largest rectangular area within the binary mask. Initially, we construct a histogram for each row of the matrix, with each entry in the histogram representing the cumulative height of masked areas in the column up to that row. We then calculate the maximum area rectangle that can be formed in each histogram, updating the coordinates of the largest rectangle as we iterate through the rows. This process ultimately yields the top-left and bottom-right coordinates of the largest rectangle fitting inside the mask. An example is shown in Fig. \ref{fig:mask_shape_example}. During inference, we stitch the hint image within the largest rectangular region of the mask.

\begin{figure}[h!]
    \centering
    \includegraphics[width=\linewidth]{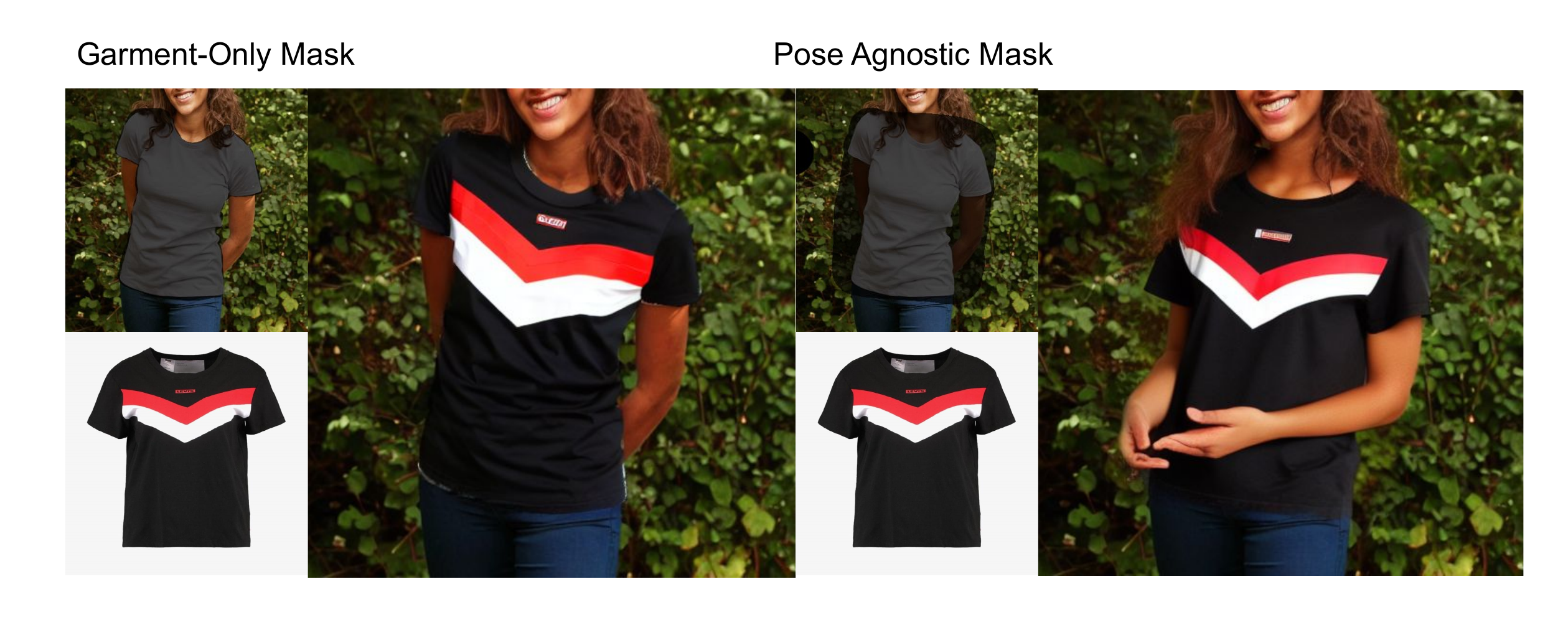}
    \caption{Pose agnostic masking case.}
    \label{fig:poseagnosticmask}
\end{figure}

\subsection{Implementation Details and Inference Performance}
In all our experiments, we used Stable Diffusion v1-5 \cite{rombach2022high}. For our image encoder, we employed DINOV2 \cite{oquab2023dinov2}, which outputs $1536$-dimensional vectors for every patch of the reference image, of shape $224 \times 224 \times 3$. Thus, it yields $256 \times 1536$-dimensional outputs. Additionally, we appended the CLS token to obtain $257 \times 1536$ image conditioning vectors. Subsequently, these vectors were processed through a single layer of a fully connected network, which was trained from scratch, to reduce them to $257 \times 768$ dimensions. We trained our model using AdamW \cite{loshchilov2017decoupled} with a constant learning rate of $1e-5$ and used horizontal flip and rotation as augmentations. To calculate the CLIP score, we used ViT-B/32 \cite{radford2021learning}. Finally, the model is efficient in inference, taking $\approx$ 6 seconds to run a single pass on an A100 (40GB) GPU with 40 DDIM steps.

\subsection{The effect of masking}

\begin{figure*}[t]
\begin{center}
  \includegraphics[width=1.0\linewidth]{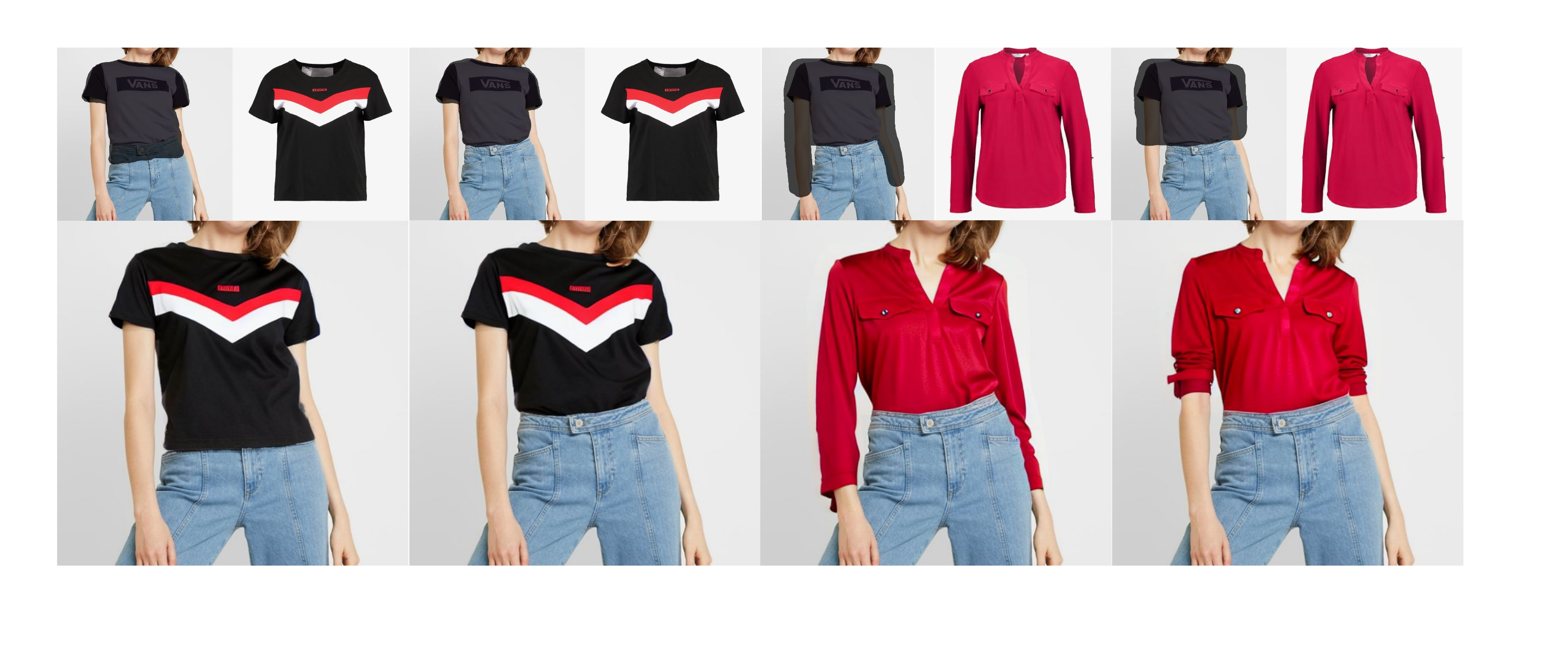}
\end{center}
  \caption{DTC allows users to manipulate different styles of the same clothing by adjusting the mask (given in the row above for each image). The first two columns display variations of the same t-shirt, showcasing it both tucked out and tucked in. The third and fourth columns illustrate the same shirt with normal sleeves and with sleeves rolled up.}
\label{fig:mask_manipulation}
\end{figure*}

Since our approach relies on collages, the mask serves as a strong prior for the DTC model. As illustrated in Fig. \ref{fig:mask_manipulation}, the use of masking enables users during inference to manipulate clothing styles. Consequently, users can guide the model to generate a t-shirt in a tucked-in style, or with sleeves rolled up, among other variations.

\subsection{Iterative Inpainting}
\begin{figure*}[t]
\begin{center}
  \includegraphics[width=1.0\linewidth]{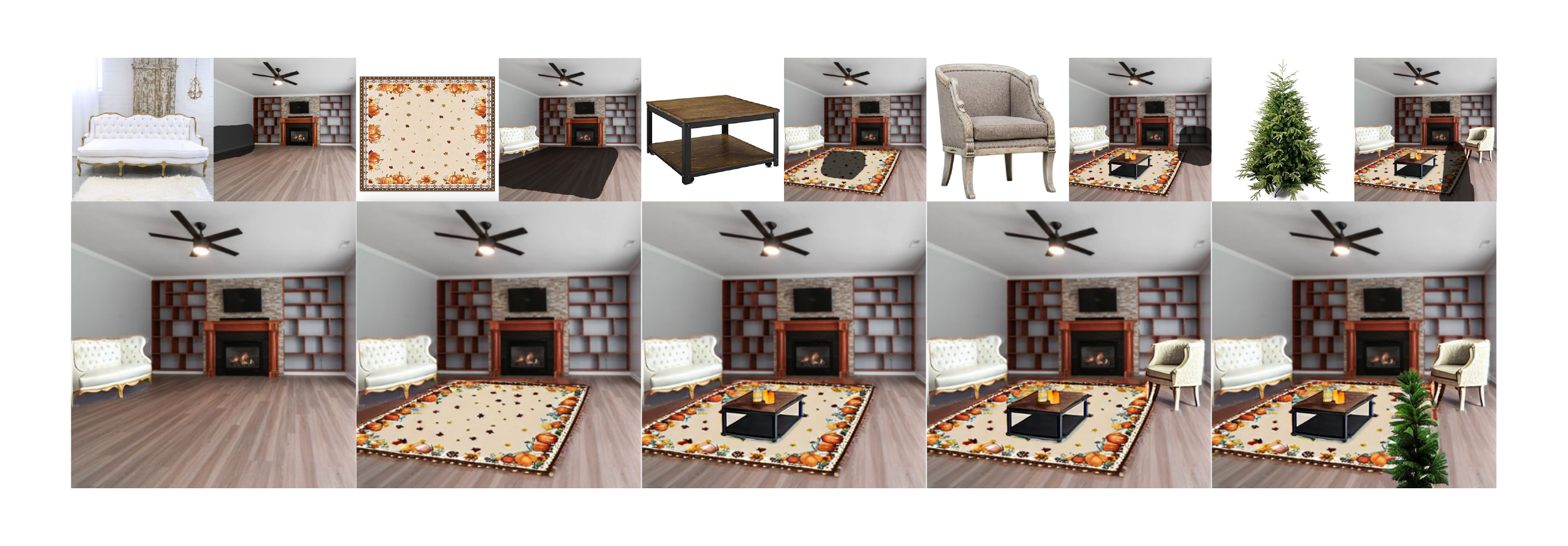}
\end{center}
  \caption{Diffuse to Choose allows users to iteratively decorate an empty room from scratch.}
\label{fig:room_decoration}
\end{figure*}

\begin{figure*}[t]
\begin{center}
  \includegraphics[width=1.0\linewidth]{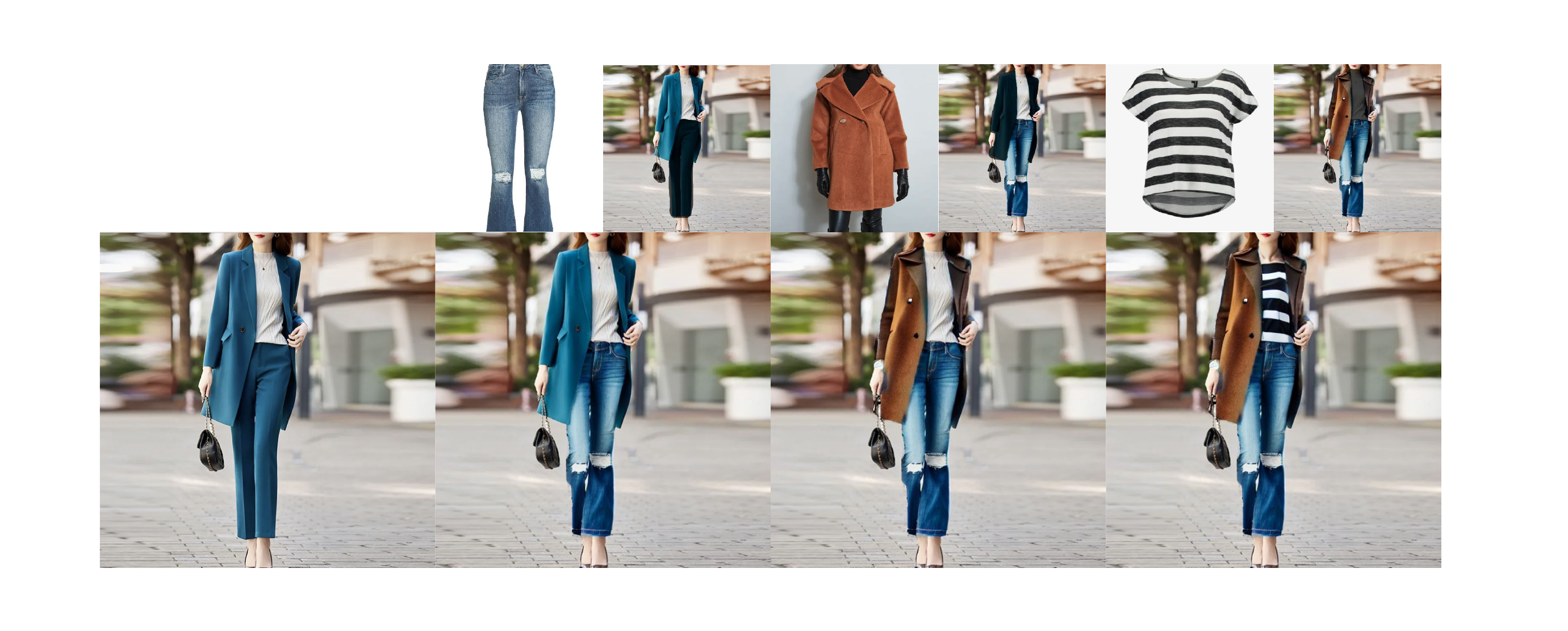}
\end{center}
  \caption{Diffuse to Choose allows users to iteratively try out combination of clothes.}
\label{fig:iterative_clothing}
\end{figure*}

DTC enables a range of enjoyable applications. For instance, users can begin with an empty room and iteratively decorate it, designing as shown in Fig. \ref{fig:room_decoration}. The same principle applies to clothing; users can generate multiple items of clothing in combination with one another to experiment with different outfit combinations, as shown in Fig. \ref{fig:iterative_clothing}.

\begin{figure*}[t]
\begin{center}
  \includegraphics[width=1.0\linewidth]{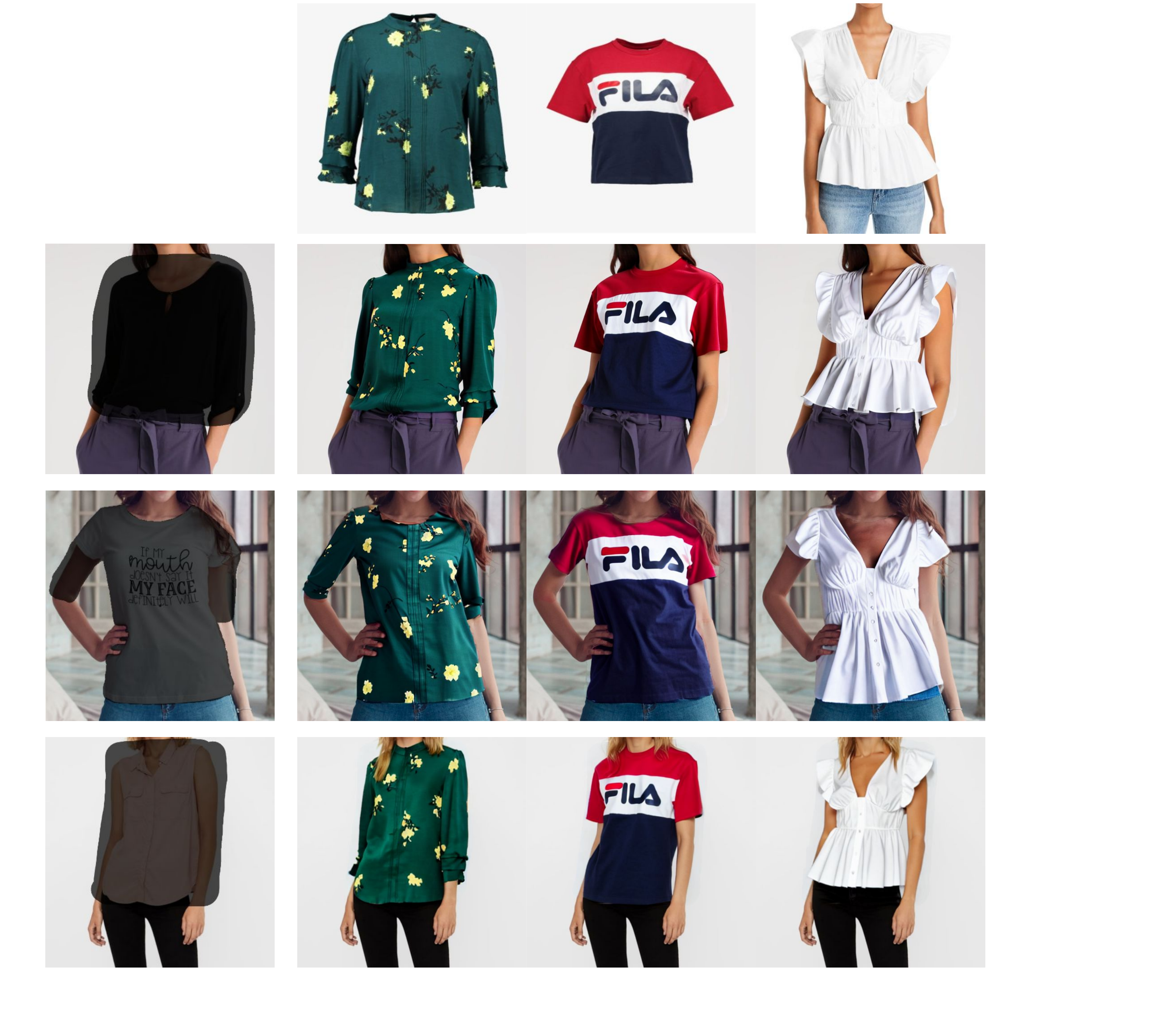}
\end{center}
  \caption{Some clothing try-on results. Note that DTC can handle in the wild reference and source images.}
\label{fig:clothing_matrix}
\end{figure*}

\subsection{Visualization of Hint Signal}

\begin{figure*}[h]
\begin{center}
  \includegraphics[width=1.0\linewidth]{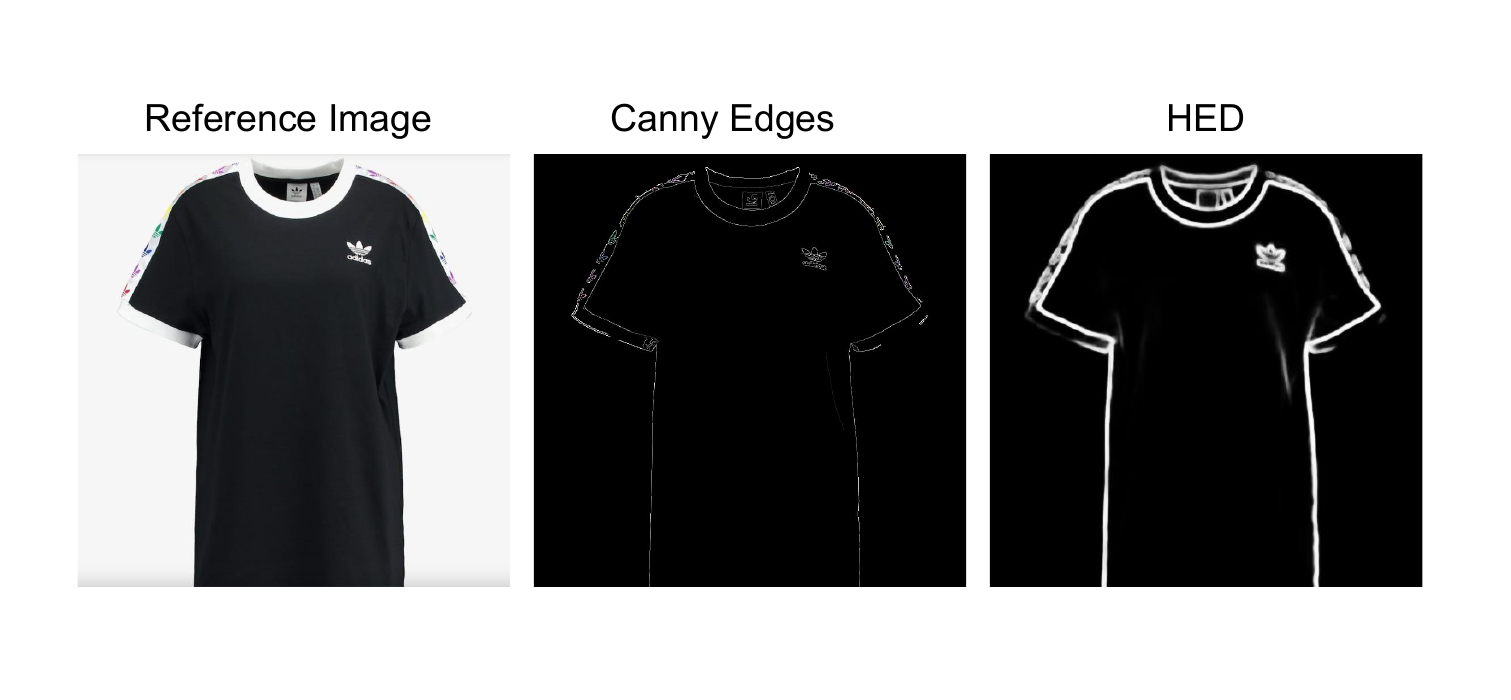}
\end{center}
  \caption{Different reference image representations that we use on hint pathway.}
\label{fig:cannyhed}
\end{figure*}

As mentioned, in addition to direct stitching, we also utilized Canny Edges and HED edges on our hint pathway, as demonstrated in Fig. \ref{fig:cannyhed}. For Canny Edges, we used sobel filters for each color channel independently and then combined the results to obtain RGB edge information, which we believed could more effectively convey the details of e-commerce items.

\subsection{More on Limitations}

For certain items, such as shoes, the model frequently fails to generate satisfactory results. We argue that this issue stems from SAM's \cite{kirillov2023segment} inability to generate appropriate masks specifically for shoes or, more broadly, for items presented in pairs. SAM often masks only one shoe of a pair, leading the model to learn shortcut features from the unmasked shoe during training, rather than acquiring useful, generalizable features. And as mentioned, since we use a latent diffusion model as our backbone, no matter how much extra information we guide it with, we are subject to the capacity of VAE decoder, which often fails to generate very fine grained concepts like detailed engravings etc. 

\subsection{Comparison Against Other Methods.}

Most state-of-the-art GAN-based methods are tailored for single-domain applications, such as virtual try-ons in controlled environments with sanitized backgrounds, and often necessitate additional inputs like pose or depth maps. Also, it is already established that diffusion-based approaches are superior to GANs in performance, possessing more comprehensive world models \cite{zhan2023does}. Consequently, diffusion-based models are more apt for the Vit-All use case.

The scope of our comparison models is intentionally limited. Personalization models such as ELITE \cite{wei2023elite}, Custom Diffusion \cite{kumari2023multi}, DreamBooth \cite{ruiz2023dreambooth}, and Textual Inversion \cite{gal2022image} lack inpainting capabilities, as they aim to directly generate entire views. DreamPaint \cite{seyfioglu2023dreampaint} is the only exception with inpainting support. Among the models that facilitate inpainting, including PBE \cite{yang2023paint} and DreamPaint, we attempted to employ DCCF \cite{xue2022dccf}. However, its tendency to create copy-paste artifacts made it unsuitable for the Vit-All task, where semantically blending the item with its environment is as crucial as preserving its detailed features.




\subsection{More Examples}

\begin{figure*}[t]
\begin{center}
  \includegraphics[width=1.0\linewidth]{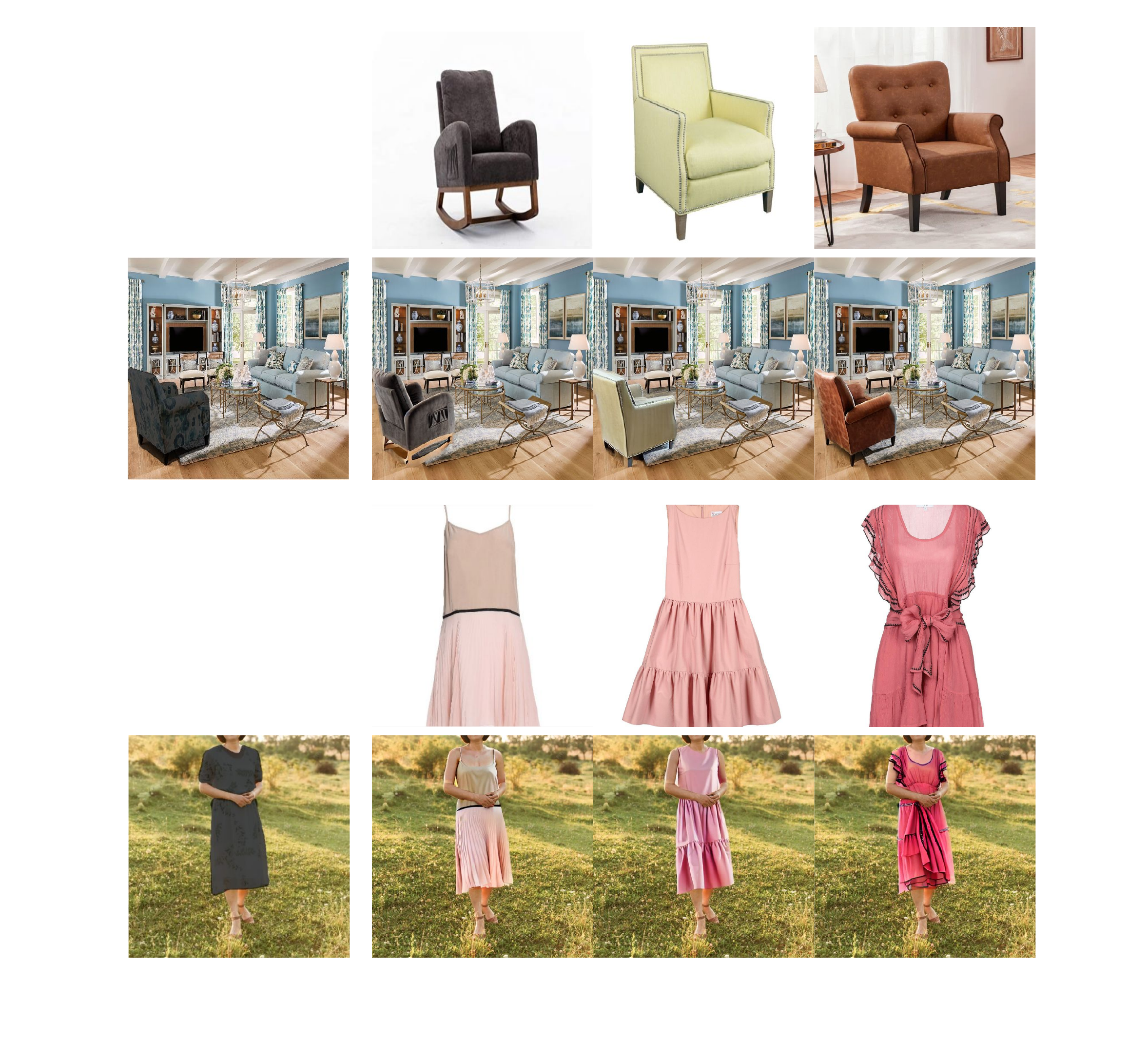}
\end{center}
  \caption{Additional examples showcasing different products. Note that DTC can infer how the product should look like, given a zero-shot example.}
\label{fig:largeexamples1}
\end{figure*}

\begin{figure*}[t]
\begin{center}
  \includegraphics[width=1.0\linewidth]{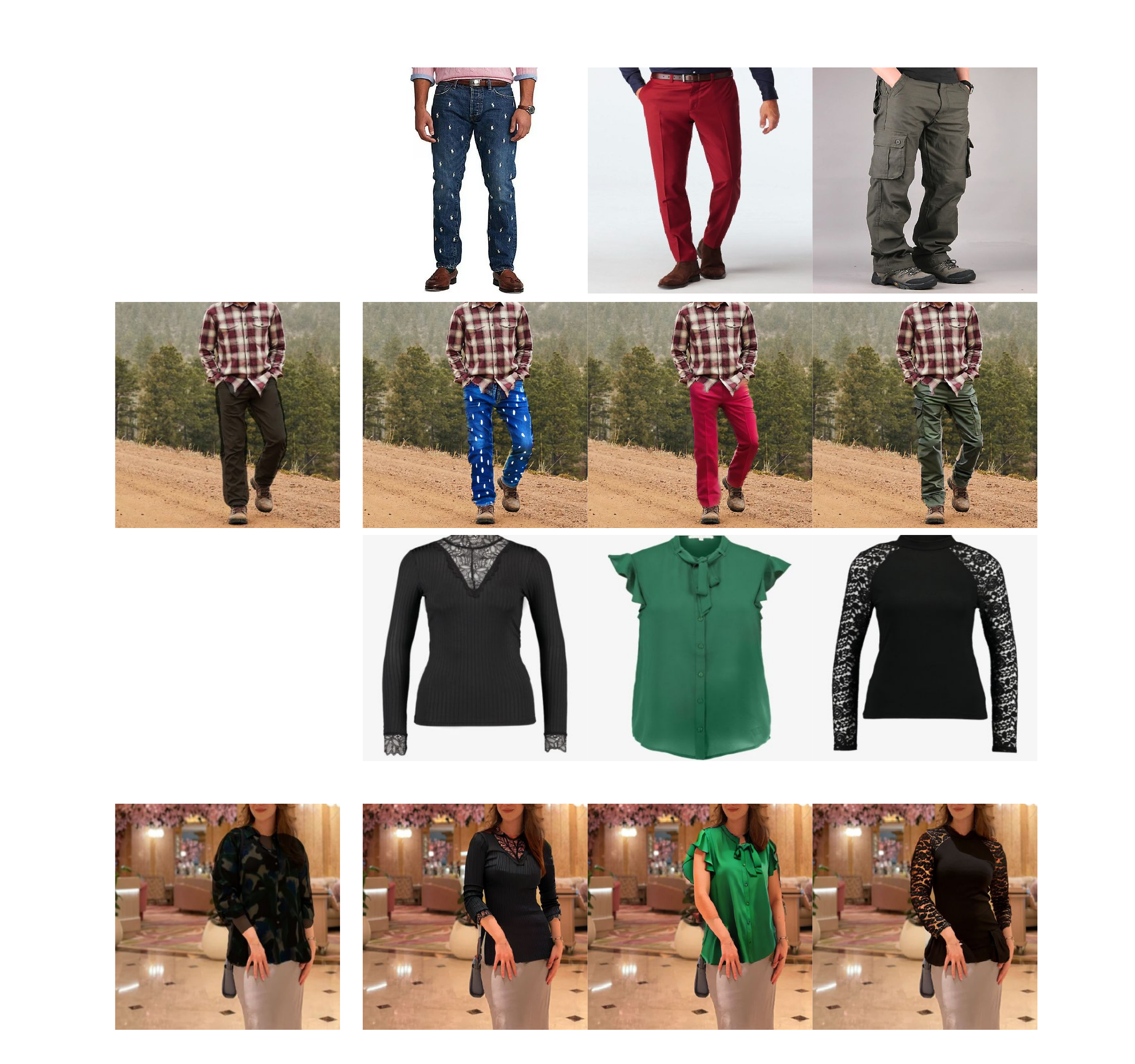}
\end{center}
  \caption{Additional examples showcasing different products.}
\label{fig:largeexamples2}
\end{figure*}

We provide more qualitative examples to showcase DTC's capabilities. Please see Fig. \ref{fig:largeexamples1}, \ref{fig:largeexamples2}, \ref{fig:clothing_matrix} for more examples.

\end{document}